\pdfoutput=1
\documentclass[11pt]{article}

\usepackage{EACL2023}

\usepackage{times}
\usepackage{latexsym}
\usepackage{amsmath}
\usepackage{booktabs}
\usepackage{xspace}

\usepackage{float}
\usepackage{bm}
\usepackage{graphicx}
\usepackage[T1]{fontenc}
\usepackage[utf8]{inputenc}

\usepackage{microtype}

\usepackage{inconsolata}

\title{Contextual Dynamic Prompting for Response Generation in Task-oriented Dialog Systems}

\author{Sandesh Swamy   \\ AWS AI Labs \\ sanswamy@amazon.com \And
        Narges Tabari \\ AWS AI Labs \\ nargesam@amazon.com
        \AND
        Chacha Chen\thanks{\quad \hspace{-1mm}Work done during an internship at AWS AI Labs}\\ University of Chicago \\ chacha@uchicago.edu \And
        Rashmi Gangadharaiah \\ AWS AI Labs \\ rgangad@amazon.com}

\begin{document}
\maketitle
\begin{abstract}
Response generation is one of the critical components in task-oriented dialog systems. 
Existing studies have shown that large pre-trained language models can be adapted to this task. The typical paradigm of adapting such extremely large language models would be by fine-tuning on the downstream tasks which is not only time-consuming but also involves significant resources and access to fine-tuning data. %However, fine-tuning billions of parameters is not only time-consuming but also involves significant use of energy and resources. 
%However, fine-tuning billions of parameters is not only time-consuming but also involves significant resources. 
%With the development of GPT-3, 
Prompting \citep{schick2020exploiting} has been an alternative to fine-tuning in many NLP tasks. In our work, we explore the idea of using prompting for response generation in task-oriented dialog systems. 
%We first explore naively applying prompting to response generation task using the prefix tuning framework. 
Specifically, we propose an approach that performs \textit{contextual dynamic prompting} where the prompts are learnt from dialog contexts. %With our approach, 
We aim to distill useful prompting signals from the dialog context. On experiments with MultiWOZ 2.2 dataset \cite{zang2020multiwoz}, we show that contextual dynamic prompts improve response generation in terms of \textit{combined score} \cite{mehri-etal-2019-structured} by 3 absolute points, and a massive 20 points when dialog states are incorporated. Furthermore, human annotation on these conversations found that agents which incorporate context were preferred over agents with vanilla prefix-tuning.
\end{abstract}

\section{Introduction}

With the advent of large language models (LLMs), a vast majority of NLP tasks, including dialog systems, further fine-tune these LMs for their downstream tasks. Although these approaches provide substantial improvements over traditional task-specific models  \cite{ham2020end,hosseini2020simple,He_Dai_Zheng_Wu_Cao_Liu_Jiang_Yang_Huang_Si_Sun_Li_2022}, it is a time consuming process that also involves significant use of energy/resources in the form of compute. These approaches also require tuning and storing parameters for each downstream task.

A more recent line of work, explores “prompting” LLMs to elicit the necessary knowledge required for the downstream tasks \cite{shin2020autoprompt,gao2020making,schick2020exploiting,petroni2019language,lee-etal-2021-dialogue,zhu2022continual}. Prompts composed of tokens or short pieces of text (\textit {discrete} prompts) inserted at the end of the input examples. These prompts are typically manually defined based on the specific downstream task. 
%Consider a sentiment classification task, for an input sentence such as \textit{the movie was so realistic}, a template such as \textit{It was \_\_} can be appended to the input sentence. The token predicted at the end of the sentence (for example, \textit{It was \underline{great}}) can then be used to map to a positive or a negative sentiment label. These approaches require no new parameters and are useful in few-shot scenarios where only a handful number of examples are available for fine-tuning. 
The main motivation behind these approaches stems from the idea that the large corpora that these language models are trained on contain relevant information which is pertinent to the task on hand. %Prompts help to draw out that information from the pre-trained language models (PLMs).

\textit{Adapter-tuning} was proposed as an alternate approach to fine-tuning. These methods only train task-specific layers that are inserted within pretrained LMs. Such a lightweight approach that add about 4\% task-specific parameters has shown to obtain comparable performances to their fine-tuning counterparts \cite{Rebuffi:17,Houlsby:19,lin-etal-2020}. 

Drawing inspiration from prompting, \textit{prefix-tuning} approaches \cite{li2021prefix} were proposed as another alternative to fine-tuning. These approaches pre-pend a sequence of task-specific \textit{continuous} vectors (aka prefix-) to the input. In contrast to prompting, the prefix consists of free parameters that do not correspond to actual real tokens. Such an approach is more prevalent since it only optimizes the prefix and does not tune parameters of the entire LM. 

Most of the existing approaches use static prompts, i.e., the same set of tokens are used as ``prompt tokens" regardless of input. However, we believe that taking context into consideration is critical especially in response generation since the current response has to fit not only the domain but also the information being requested in previous turns. For example: In the MultiWOZ dataset, if a customer asks about train bookings, the agent response has to restrict itself to that particular domain. To address this problem, we explore the idea of generating input-dependent or \textit{contextual} prompts. We want the prompts to capture and encode different signals for different turns of dialogs depending on the context, hence, we call our approach \textit{dynamic context prompting}. This way, we hope to distill useful signals into the prompts and provide the model with adequate signals to generate a desired system response. 
In this work, we explore the potential of using dialog context within a prefix tuning approach for the task of response generation in task-oriented dialog systems (TOD).
The contributions of this paper are summarized as:
\begin{itemize}
    \item we propose a context-dependent prefix-tuning method for dialog response generation in TOD systems.
    \item to illustrate the benefits of such an approach, we conduct experiments on the MultiWOZ dataset. We show that our model significantly outperforms the original task-dependent design of the prefix-tuning method. %Moreover, we also achieve comparable performance on adaptor and fine-tuning method with significantly less parameters.
\end{itemize}

\section{Related Work}

\subsection{Dialog Generation}
With the prevalence of LLMs, the quest for an answer to ``how do we effectively adapt such models for dialog generation?" has been on the forefront of researchers' minds in the dialog community. 
For task-oriented dialogs, fine-tuning large pre-trained models such as GPT-2 or T5 has made great progress on benchmarks recently~\cite{ham2020end,hosseini2020simple}. Built upon these advances, more recent line of work investigates the effectiveness of using multi-task learning~\cite{su2021multi,lin2020mintl,yang2021ubar}, or pre-training the model on external dialog corpora~\cite{peng2021soloist,liu2021pretraining}.
More recently, prompting has been used to address the sub-task of dialog state tracking~\cite{lee-etal-2021-dialogue,zhu2022continual}. Different from those works, we focus on the task of dialog response generation.
% There are a few works~\cite{zhu2022continual} explore the idea of prompting in task-oriented dialog systems.

\subsection{Prompt-based Learning}
As an alternative to the fine-tuning paradigm, prompting involves a sequence of tokens appended to the input text, which can then induce the model to engage in a certain behavior suited to the task. %For example, in the GPT-2 paper~\cite{radford2019language}, ``tl;dr:'' is used to prompt the model to generate summarizing texts. 
% For example, in machine translation, we can use prompts such as ``translate to French, [X], [Z]'' where [X] is the input example and [Z] is the desired output text. 
Since the release of GPT-2~\cite{radford2018improving,radford2019language,brown2020language}, many prompt-related papers have emerged. Most of the leading approaches in prompting use \textbf{task-specific prompts}, ranging from discrete prompts~\cite{shin2020autoprompt,gao2020making,schick2020exploiting,petroni2019language} to continuous ``soft prompts''~\cite{li2021prefix,lester2021power}. %Those methods usually work in a transfer learning setting, where the prompt is fixed for each task.
These methods have a fixed prompt for each task.
However, in dialog systems specifically, the context varies for every turn. In our work, we aim to design prompts which are \textbf{context-dependent}.

\section{Problem Statement}
\textbf{Response generation} is one of the tasks carried out in dialog systems usually in addition to dialog state tracking (DST). Given a dialog context (previous turns between the system and the user) $C = [u_1,s_1,...,u_{n-1}, s_{n-1}]$ and the current user utterance $u_n$, the goal of response generation is to generate system response $s_n$. Note that in the actual task, we generate \textit{delexicalized} system responses, given all the groundtruth previous turns as input, following previous works~\cite{hosseini2020simple,wen2015semantically}. 

Techniques mentioned in \cite{ham2020end, hosseini2020simple} rely on fully fine-tuning LLMs to carry out this task. In contrast, our approach builds on the prefix-tuning framework, but incorporates dialog context, $C$, as an additional signal for the prefix tokens. As a supplement to context $C$, we added dialog state information  $D$ (up to the current turn) to further help response generation.

\section{Contextual Dynamic Prompting Framework}
\label{Conxtual dynamic prompting}
% \subsection{T5 for Response Generation}

%In this section, we start by adapting prefix tuning for response generation. Then, we introduce our design of contextual dynamic prompting encoder. 

\subsection{Prefix-tuning for Response Generation}
\label{sec:prefix-tuning}

\begin{figure*}
    \vspace{-3mm}
    \centering
    \includegraphics[width=0.8\textwidth, height=30mm]{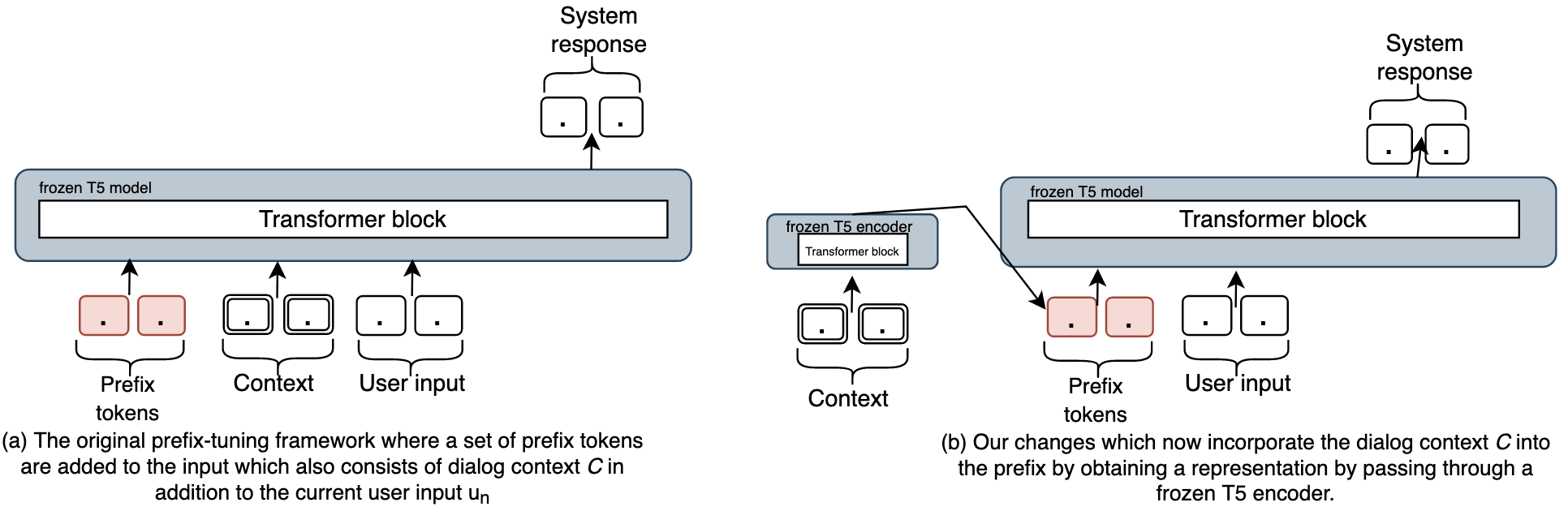}
    \caption{\small The figures above indicate the differences between the vanilla prefix-tuning approach compared to our approach. In both these variants, only the prefix tokens are tuned.}
    \label{fig:contextual dynamic prompting}
\end{figure*}

Our work is built on top of prefix tuning for generation tasks~\cite{li2021prefix}, which adds a fixed set of tunable prefix tokens/prompts to the original input $x$ to obtain a new input, [PREFIX; $x$]. Following the denotation in \citep{li2021prefix}, we use $P_\theta[i,:]$ to denote the $i_{th}$ prefix. $P_\theta[i,:]$ is generated by:
\begin{equation}
\label{eq:mlp}
    P_\theta[:,:] = MLP_\theta(P'),
\end{equation}
where $P'$ is a fixed smaller matrix as input to a feedforward neural network ($MLP_\theta$). The training objective of prefix-tuning is same as fine-tuning, i.e., the following log-likelihood objective:
\begin{equation*}
    \max_\theta \log p_\phi (y|x),
\end{equation*}
where $y$ is the decoder output and $x$ is the input. $\theta$ represents the trainable parameters in the prefix tuning feedforward neural network and $\phi$ denotes all other parameters that include the frozen parameters of the large language model. 

For our task of response generation, we concatenate the prefix with the dialog context and the current user utterance as input [PREFIX; $u_1,s_1,...,u_{n-1}, s_{n-1},u_n$]. The target output is the system response $s_n$ as seen in Figure \ref{fig:contextual dynamic prompting} (a).

We adopt T5~\cite{raffel2020exploring} as the pre-trained language model. T5 employs an encoder-decoder framework which is prevalent in seq2seq tasks \cite{sutskever-s2s, cho-etal-2014-properties}.

\subsection{Contextual Prefix-tuning}
In vanilla prefix-tuning, the parameters of the prefix are fixed after training for any particular task to be reused. However, a dialog system involves having multiple turns of conversation between a system and the user. It is imperative in such systems to dynamically incorporate contextual information to carry out a meaningful conversation with the user. We explore how we can distill the dialog context information into the prefix with a prompt encoder. 

Different from the original design, we want to encode additional signals into the prefix that differs for each input instances. In other words, we want to generate \textbf{contextual prefix} or \textbf{contextual dynamic prompts}. 

Formally, we modify the equation~(\ref{eq:mlp}) as follows:
\begin{equation}
\label{eq:mlp_cdp}
    P_\theta[:,:] = MLP_\theta(encoder(C)),
\end{equation}
where  $C = [u_1,s_1,...,u_{n-1}, s_{n-1}]$ represents the dialog context.
We first obtain the representation of the dialog context by feeding $C$ into a T5 encoder which is kept frozen as shown in Figure \ref{fig:contextual dynamic prompting} (b). Subsequently, we use the prompt encoder, i.e., the feedforward neural network, to get the prefix. 
The generated prefix $P_\theta$ is then concatenated with only the current user utterance. Instead of concatenating the whole context as the input to the T5 decoder, we first distill the signal into the prefix tokens. As a consequence of freezing the T5 encoder which generates the context representation, we still have the same number of tunable parameters as the original prefix-tuning framework.

\subsection{Input-dependent Prefix-tuning with Dialog State}
In most task-oriented dialog systems, we also have access to the dialog state at every turn in addition to dialog context. The dialog state has information such as requested slots and filled slots at every turn. We provide the dialog state $D$ in addition to the context $C$ to obtain contextual dynamic prompts. As a result, we will now modify  equation~(\ref{eq:mlp_cdp}) as:
\begin{equation}
    \label{eq:mlp_cdp_ds}
    P_\theta[:,:] = MLP_\theta(encoder(C;D_{n-1})),
\end{equation}

we only provide the most recent dialog state $D_{n-1}$ which is an amalgamation of all previous dialog states $D_{<n-1}$.

\begin{table*}[ht!]
\centering
\vspace{-0.1in}
\resizebox{\textwidth}{!}{
\begin{tabular}{lcccccccc} 
\toprule 
& \multicolumn{8}{c}{MultiWOZ 2.2}  \\
            &   BLEU & Inform  & Success   & Combined Score & Av. len. & \#uniq. words	 & \#uniq. 3-grams \\
\midrule
% FINE-TUNE    & 20.245 & 75.7 & 67.4 & 91.795 &  13.957 & 1.828 & 371 & 3084 \\
% FINE-TUNE (with states) &        \\
% ADAPTOR &  19.4524 & 77.6 &  68.6 & 92.5525 & 13.5457\\
% ADAPTOR (with states) & 20.248 & 73.7 &  67.9 &  91.048 & 14.448 & 1.693 & 299 & 2400 & \\
Prefix-Tuning & 19.19 & 54.7 & 48.0 & 70.54 & 13.83 & 245 & 1671 & \\
Prefix-Tuning (with DS) & 19.36 & 51.8 &  47.0 & 68.76 & 13.08 & 231 & 1626 \\
% \midrule
Contextual Dynamic Prompt  & 19.16 & 58.1 &  50.5 & 73.46 & \textbf{14.16} & 231 & 1532 \\
Contextual Dynamic Prompt (with DS)  & 17.94 & \textbf{77.2}  &  \textbf{68.8} & \bf  90.94 & 14.02 & \textbf{282} & \textbf{2390} \\
% \midrule
% SOTA-external sdata\\
% \midrule
% SOTA (GALAXY) & 19.64 & 85.4 & 75.7 & 100.2 & 13.39 & 1.75& 295 & 2275\\
\bottomrule 
\end{tabular}
}
\caption{\textbf{Performance Comparison}. All model performance are based on features from all modalities. Contextual Dynamic Prompt (with DS) has the best performance in combined score. %thanks to its  efficient data fusion module that captures individual-level features, and the uncertainty modeling that learns population-level correlation.
}
\label{tab:all_result}
\end{table*}

\section{Experimental Settings}

\subsection{Dataset and Metrics}

We evaluate our proposed framework and model on the MultiWOZ 2.2 dataset~\cite{zang2020multiwoz,budzianowski2018large} which is a large-scale, multi-domain, human-human task-oriented dialog dataset collected via the
Wizard-of-Oz framework where one participant plays the
role of the system. It consists of seven domains including hotel, restaurant, attraction, train, taxi, hospital, and police, and an additional
domain general for acts such as greeting or goodbye. 
Due to its multi-domain setting, complex ontology, and flexible human expressions, developing dialog systems on MultiWOZ is extremely challenging. 
The training data contain 8437 dialogs, the dev and test set contain 1000 dialogs each. 

% \begin{table}[htbp]
% \caption{Data statistics of MultiWOZ 2.2.} 
% % \vspace{-0.1in}
% \label{tab:ehr_statistics}
% \centering 
% \begin{tabular}{lccc}
% \toprule
%               &   Train &   Dev &  Test       \\ 
%     \midrule
% % Prevalance & 18.75\% &        .                   \\ 
% \# Dialogs & 8437 & 1000 & 1000 \\
% % \# Dev & 1000\\
% % \# Test & 1000
% \bottomrule
% \end{tabular} 
% \end{table}

We use four evaluation metrics: \textbf{BLEU}~\cite{papineni2002bleu}, \textbf{Inform}, and \textbf{Success} rates, and \textbf{combined score}. \textit{Inform} measures whether the system provides an appropriate entity and \textit{Success} measures whether the system answers all the requested attributes. Specifically, the Inform rate relates to attributes that allow the user to constrain database searches, e.g., restaurant location or price range (the informational slots) and the Success rate focuses on request-able slots, that can be asked by the user, e.g., phone number. Both are calculated on the level of dialogs. The combined score is calculated following \cite{mehri2019structured} as \( BLEU + 0.5 * (Inform + Success) \). We followed a standard script~\footnote{\url{https://github.com/Tomiinek/MultiWOZ_Evaluation}} to report different measures.

\subsection{Human Evaluation}
\label{section:human_eval}
%We randomly chose 10\% of conversations form the final evaluation set with the total of 728 turns across 100 conversations. Each annotator was tasked to pick the best turn out of the 4 different Agents. The agent numbers, when provided to annotators, were shuffled to avoid bias. Agents are described as follows:
We chose a 10\% subset of the evaluation set (randomly shuffled) conversations with a total of 728 turns across them and provided annotators with the responses generated by each of the methods described in section \ref{Conxtual dynamic prompting}. Annotators were asked to rate each agent on a turn-level and to also pick the agent which carried out the best conversation. If annotators felt more than one agent did well, they could choose multiple agents. The agent numbers, when provided to annotators, were shuffled to avoid bias. Each agent is described as:
\begin{itemize}
    \itemsep0em 
    \item{Agent 1}: Incorporates only prefix-tuning 
    \item{Agent 2}: Incorporates prefix-tuning with Dialog State
    \item{Agent 3}: Incorporates contextual dynamic prompts
    \item{Agent 4}: Incorporates contextual dynamic prompts with Dialog State
\end{itemize}

When annotating on turn level, from these 728 turns, we saw that the agents tied on 596 occasions, agent 1 had outright win on 12 occasions, agent 2 on 22, agent 3 on 33 occasions, and agent 4 on 65 occasions. This shows that our technique of using contextual dynamic prompts for generating responses is effective (Examples in Appendix \ref{appendix:eval}). 
%The full table/figure for turn level annotation can be found in Appendix. 

%We also asked our annotators to perform a conversation level annotation where they were asked to pick the agent which had the best overall response to the user.  Once again,
Additionally, on the conversation level, we noticed that across 100 conversations, 37 were tied, and agents 3 and 4 were preferred in a total of \textit{53} conversations confirming our hypothesis that incorporating context into prompts leads to better responses. We request readers to refer to Appendix \ref{appendix:ex} and \ref{appendix:eval} for more details about the annotation task.

\section{Results}

As shown in Table \ref{tab:all_result}, contextual dynamic prompting with dialog states obtains a combined score of \textit{90.94}, a 20 point jump from our baseline (prefix-tuning). In addition, even though we can't explicitly explain the drop in BLEU, the massive jumps in both success and inform suggest more transparency and coherence for the responses generated by the input-dependent prefix-tuning as these metrics focus on quality of informational and request-able slots in each turn. When comparing our results with the human annotations, we also see that \textit{Agent 4} - which uses contextual dynamic prompting - wins 38 conversations (out of 100). This is majority of wins compared to \textit{Agent 1} winning only 3 conversations, and \textit{Agent 2} winning 7. This further emphasized that adding contextual dynamic prompts leads to better quality of responses. We show some examples from the human annotation task where \textit{Agent 4} was chosen as the winning agent in Tables \ref{tab:example_1} and \ref{tab:example_3} below.

\begin{table*}[!th]
    \small
  \centering
  \begin{tabular}{|c|p{25mm}|p{25mm}|p{25mm}|p{25mm}|p{25mm}|}
    \hline
    Turn num & User turn & Agent 1 response & Agent 2 response & Agent 3 response & Agent 4 response \\
    \hline
    1 & i am looking for a place to dine. the restaurant should be expensive and should serve chinese food. & there are [choice] [pricerange] [food] restaurants in town. do you have a preference for area or price range? & i have [choice] [pricerange] [food] restaurants. do you have a preference for area? & there are [choice] [pricerange] [food] restaurants. do you have a price range in mind? & there are [choice] [food] restaurants in the [pricerange] price range. is there a particular area you are interested in?\\
    \hline
    2 & not in any specific area please. i need to book a table for 3 at 14:00 on saturday.  & booking was successful. the table will be reserved for 15 minutes. reference number is : [ref]. is there anything else i can help you with?& booking was successful. the table will be reserved for 15 minutes. reference number is : [ref].& i was able to book you at [name] for [bookday] at [booktime]. your reference number is [ref]. is there anything else i can help you with?& i have made a reservation for you at [name]. the table will be reserved for 15 minutes. the reference number is [ref]. is there anything else i can help you with?\\
    \hline
    3& how about 13:00 then? & booking was successful. the table will be reserved for 15 minutes. reference number is : [ref]. & booking was successful. the table will be reserved for 15 minutes. reference number is : [ref]. & booking was successful. the table will be reserved for 15 minutes. reference number is : [ref]. is there anything else i can help you with? & ok, i have you booked at [name]. your reference number is [ref]. is there anything else i can help you with? \\
    \hline
    4& what is the reference number for my booking?  & your reference number is [ref]. is there anything else i can help you with? & the reference number is [ref]. is there anything else i can help you with? & your reference number is [ref]. is there anything else i can help you with? & the reference number is [ref].\\
    \hline
    5& no thank you, that is all i needed!  & you're welcome. have a great day! & you're welcome. have a great day! & you're welcome. have a great day!& you're welcome, have a great day! \\
    \hline
  \end{tabular}
  \caption{An example where Agent 4 was picked as the best performing agent in a conversation.}
  \label{tab:example_1}
\end{table*}

\begin{table*}[!th]
    \small
  \centering
  \begin{tabular}{|c|p{25mm}|p{25mm}|p{25mm}|p{25mm}|p{25mm}|}
    \hline
    Turn num & User turn & Agent 1 response & Agent 2 response & Agent 3 response & Agent 4 response \\
    \hline
    1 & can you help me find a restaurant that serves african food with a moderate price range please  & there are no [food] restaurants in the [pricerange] price range. would you like to try a different type of food? & there are no [food] restaurants in the [pricerange] price range. would you like to try a different type of food? & there are no [food] restaurants in the [pricerange] price range. would you like to try a different type of food? & [name] is a [food] restaurant in the [area]. would you like to try it?\\
    \hline
    2 & yes how about asian food in the same price range?  & there are [choice] [food] restaurants in the [area] and [choice] in the [area]. do you have a preference?& there are [choice] [food] restaurants in the [pricerange] price range. do you have a preference for area?& there are [choice] [pricerange] [food] restaurants in the [area]. do you have a price range in mind?& [name] is a [food] restaurant in the [pricerange] price range.\\
    \hline
    3& will you please book a table for 6 at 13:45 on friday?  & booking was successful. the table will be reserved for 15 minutes. reference number is : [ref]. is there anything else i can help you with? & booking was successful. the table will be reserved for 15 minutes. reference number is : [ref]. & i was able to book you at [name] for [bookday] at [booktime]. your reference number is [ref]. is there anything else i can help you with? & i have made a reservation for [bookpeople] at [booktime] on [bookday] at [booktime]. the table will be reserved for 15 minutes. the reference number is [ref]. is there anything else i can help you with? \\
    \hline
    4& that's all i needed. thanks a lot for the help! & you're welcome. have a great day! & you're welcome. have a great day! & you're welcome. have a great day! & you're welcome. have a great day!\\
    \hline
  \end{tabular}
  \caption{An example where Agent 4 was picked as the best performing agent in a conversation.}
  \label{tab:example_3}
\end{table*}

\section{Conclusion}
In our work, we proposed an approach that performs \textit{contextual dynamic prompting} where prompts are learnt from dialog contexts with the goal of distilling useful prompting signals. In our experiments, we showed that contextual dynamic prompts improve response generation in terms of \textit{combined score} \cite{mehri-etal-2019-structured} by 3 points, and by 20 points when \textit{dialog states} are incorporated compared to the baseline. Our technique does not expose the models to additional knowledge sources. Human annotation on these conversations found that agents which incorporate context into prompts were preferred over agents with vanilla prefix-tuning.

\section*{Limitations}
While our work explores a new technique of contextual dynamic prompts for response generation, we carried out our experiments on a dataset which is in the English language. A potential limitation of this work would be the transfer of our findings on an English dataset to a multi-lingual dataset or a mono-lingual dataset on a language other than English. We plan to address this in our future work and also request the help of the research community in doing so.

%\section*{Ethics Statement}
%Scientific work published at EACL 2023 must comply with the \href{https://www.aclweb.org/portal/content/acl-code-ethics}{ACL Ethics Policy}. We encourage all authors to include an explicit ethics statement on the broader impact of the work, or other ethical considerations after the conclusion but before the references. The ethics statement will not count toward the page limit (8 pages for long, 4 pages for short papers).

\section*{Acknowledgements}
We would like to thank the anonymous reviewers for their feedback on this work. We also thank the AWS Console/UXP2 Science team at AWS AI Labs for discussions around this work. Lastly, we'd like to thank the human annotators for their efforts.

% Entries for the entire Anthology, followed by custom entries
\bibliography{anthology,custom}

\begin{thebibliography}{31}
\expandafter\ifx\csname natexlab\endcsname\relax\def\natexlab#1{#1}\fi

\bibitem[{Brown et~al.(2020)Brown, Mann, Ryder, Subbiah, Kaplan, Dhariwal,
  Neelakantan, Shyam, Sastry, Askell et~al.}]{brown2020language}
Tom Brown, Benjamin Mann, Nick Ryder, Melanie Subbiah, Jared~D Kaplan, Prafulla
  Dhariwal, Arvind Neelakantan, Pranav Shyam, Girish Sastry, Amanda Askell,
  et~al. 2020.
\newblock Language models are few-shot learners.
\newblock \emph{Advances in neural information processing systems},
  33:1877--1901.

\bibitem[{Budzianowski et~al.(2018)Budzianowski, Wen, Tseng, Casanueva, Stefan,
  Osman, and Ga{\v{s}}i\'c}]{budzianowski2018large}
Pawe{\l} Budzianowski, Tsung-Hsien Wen, Bo-Hsiang Tseng, I{\~n}igo Casanueva,
  Ultes Stefan, Ramadan Osman, and Milica Ga{\v{s}}i\'c. 2018.
\newblock Multiwoz - a large-scale multi-domain wizard-of-oz dataset for
  task-oriented dialogue modelling.
\newblock In \emph{Proceedings of the 2018 Conference on Empirical Methods in
  Natural Language Processing (EMNLP)}.

\bibitem[{Cho et~al.(2014)Cho, van Merri{\"e}nboer, Bahdanau, and
  Bengio}]{cho-etal-2014-properties}
Kyunghyun Cho, Bart van Merri{\"e}nboer, Dzmitry Bahdanau, and Yoshua Bengio.
  2014.
\newblock \href {https://doi.org/10.3115/v1/W14-4012} {On the properties of
  neural machine translation: Encoder{--}decoder approaches}.
\newblock In \emph{Proceedings of {SSST}-8, Eighth Workshop on Syntax,
  Semantics and Structure in Statistical Translation}, pages 103--111, Doha,
  Qatar. Association for Computational Linguistics.

\bibitem[{Gao et~al.(2020)Gao, Fisch, and Chen}]{gao2020making}
Tianyu Gao, Adam Fisch, and Danqi Chen. 2020.
\newblock Making pre-trained language models better few-shot learners.
\newblock \emph{arXiv preprint arXiv:2012.15723}.

\bibitem[{Ham et~al.(2020)Ham, Lee, Jang, and Kim}]{ham2020end}
Donghoon Ham, Jeong-Gwan Lee, Youngsoo Jang, and Kee-Eung Kim. 2020.
\newblock End-to-end neural pipeline for goal-oriented dialogue systems using
  gpt-2.
\newblock In \emph{Proceedings of the 58th Annual Meeting of the Association
  for Computational Linguistics}, pages 583--592.

\bibitem[{He et~al.(2022)He, Dai, Zheng, Wu, Cao, Liu, Jiang, Yang, Huang, Si,
  Sun, and Li}]{He_Dai_Zheng_Wu_Cao_Liu_Jiang_Yang_Huang_Si_Sun_Li_2022}
Wanwei He, Yinpei Dai, Yinhe Zheng, Yuchuan Wu, Zheng Cao, Dermot Liu, Peng
  Jiang, Min Yang, Fei Huang, Luo Si, Jian Sun, and Yongbin Li. 2022.
\newblock \href {https://doi.org/10.1609/aaai.v36i10.21320} {Galaxy: A
  generative pre-trained model for task-oriented dialog with semi-supervised
  learning and explicit policy injection}.
\newblock \emph{Proceedings of the AAAI Conference on Artificial Intelligence},
  36(10):10749--10757.

\bibitem[{Hosseini-Asl et~al.(2020)Hosseini-Asl, McCann, Wu, Yavuz, and
  Socher}]{hosseini2020simple}
Ehsan Hosseini-Asl, Bryan McCann, Chien-Sheng Wu, Semih Yavuz, and Richard
  Socher. 2020.
\newblock A simple language model for task-oriented dialogue.
\newblock \emph{Advances in Neural Information Processing Systems},
  33:20179--20191.

\bibitem[{Houlsby et~al.(2019)Houlsby, Giurgiu, Jastrzebski, Morrone,
  de~Laroussilhe, Gesmundo, Attariyan, and Gelly}]{Houlsby:19}
Neil Houlsby, Andrei Giurgiu, Stanislaw Jastrzebski, Bruna Morrone, Quentin
  de~Laroussilhe, Andrea Gesmundo, Mona Attariyan, and Sylvain Gelly. 2019.
\newblock \href {http://arxiv.org/abs/1902.00751} {Parameter-efficient transfer
  learning for {NLP}}.
\newblock \emph{CoRR}, abs/1902.00751.

\bibitem[{Lee et~al.(2021)Lee, Cheng, and Ostendorf}]{lee-etal-2021-dialogue}
Chia-Hsuan Lee, Hao Cheng, and Mari Ostendorf. 2021.
\newblock \href {https://doi.org/10.18653/v1/2021.emnlp-main.404} {Dialogue
  state tracking with a language model using schema-driven prompting}.
\newblock In \emph{Proceedings of the 2021 Conference on Empirical Methods in
  Natural Language Processing}, pages 4937--4949, Online and Punta Cana,
  Dominican Republic. Association for Computational Linguistics.

\bibitem[{Lester et~al.(2021)Lester, Al-Rfou, and Constant}]{lester2021power}
Brian Lester, Rami Al-Rfou, and Noah Constant. 2021.
\newblock The power of scale for parameter-efficient prompt tuning.
\newblock \emph{arXiv preprint arXiv:2104.08691}.

\bibitem[{Li and Liang(2021)}]{li2021prefix}
Xiang~Lisa Li and Percy Liang. 2021.
\newblock Prefix-tuning: Optimizing continuous prompts for generation.
\newblock \emph{arXiv preprint arXiv:2101.00190}.

\bibitem[{Lin et~al.(2020{\natexlab{a}})Lin, Madotto, and Fung}]{lin-etal-2020}
Zhaojiang Lin, Andrea Madotto, and Pascale Fung. 2020{\natexlab{a}}.
\newblock \href {https://doi.org/10.18653/v1/2020.findings-emnlp.41} {Exploring
  versatile generative language model via parameter-efficient transfer
  learning}.
\newblock In \emph{Findings of the Association for Computational Linguistics:
  EMNLP 2020}, pages 441--459, Online. Association for Computational
  Linguistics.

\bibitem[{Lin et~al.(2020{\natexlab{b}})Lin, Madotto, Winata, and
  Fung}]{lin2020mintl}
Zhaojiang Lin, Andrea Madotto, Genta~Indra Winata, and Pascale Fung.
  2020{\natexlab{b}}.
\newblock Mintl: Minimalist transfer learning for task-oriented dialogue
  systems.
\newblock \emph{arXiv preprint arXiv:2009.12005}.

\bibitem[{Liu et~al.(2021)Liu, Yu, Rimell, and Blunsom}]{liu2021pretraining}
Qi~Liu, Lei Yu, Laura Rimell, and Phil Blunsom. 2021.
\newblock Pretraining the noisy channel model for task-oriented dialogue.
\newblock \emph{Transactions of the Association for Computational Linguistics},
  9:657--674.

\bibitem[{Mehri et~al.(2019{\natexlab{a}})Mehri, Srinivasan, and
  Eskenazi}]{mehri-etal-2019-structured}
Shikib Mehri, Tejas Srinivasan, and Maxine Eskenazi. 2019{\natexlab{a}}.
\newblock \href {https://doi.org/10.18653/v1/W19-5921} {Structured fusion
  networks for dialog}.
\newblock In \emph{Proceedings of the 20th Annual SIGdial Meeting on Discourse
  and Dialogue}, pages 165--177, Stockholm, Sweden. Association for
  Computational Linguistics.

\bibitem[{Mehri et~al.(2019{\natexlab{b}})Mehri, Srinivasan, and
  Eskenazi}]{mehri2019structured}
Shikib Mehri, Tejas Srinivasan, and Maxine Eskenazi. 2019{\natexlab{b}}.
\newblock Structured fusion networks for dialog.
\newblock \emph{arXiv preprint arXiv:1907.10016}.

\bibitem[{Papineni et~al.(2002)Papineni, Roukos, Ward, and
  Zhu}]{papineni2002bleu}
Kishore Papineni, Salim Roukos, Todd Ward, and Wei-Jing Zhu. 2002.
\newblock Bleu: a method for automatic evaluation of machine translation.
\newblock In \emph{Proceedings of the 40th annual meeting of the Association
  for Computational Linguistics}, pages 311--318.

\bibitem[{Peng et~al.(2021)Peng, Li, Li, Shayandeh, Liden, and
  Gao}]{peng2021soloist}
Baolin Peng, Chunyuan Li, Jinchao Li, Shahin Shayandeh, Lars Liden, and
  Jianfeng Gao. 2021.
\newblock Soloist: Buildingtask bots at scale with transfer learning and
  machine teaching.
\newblock \emph{Transactions of the Association for Computational Linguistics},
  9:807--824.

\bibitem[{Petroni et~al.(2019)Petroni, Rockt{\"a}schel, Lewis, Bakhtin, Wu,
  Miller, and Riedel}]{petroni2019language}
Fabio Petroni, Tim Rockt{\"a}schel, Patrick Lewis, Anton Bakhtin, Yuxiang Wu,
  Alexander~H Miller, and Sebastian Riedel. 2019.
\newblock Language models as knowledge bases?
\newblock \emph{arXiv preprint arXiv:1909.01066}.

\bibitem[{Radford et~al.(2018)Radford, Narasimhan, Salimans, Sutskever
  et~al.}]{radford2018improving}
Alec Radford, Karthik Narasimhan, Tim Salimans, Ilya Sutskever, et~al. 2018.
\newblock Improving language understanding by generative pre-training.
\newblock \emph{https://openai.com/blog/language-unsupervised/}.

\bibitem[{Radford et~al.(2019)Radford, Wu, Child, Luan, Amodei, Sutskever
  et~al.}]{radford2019language}
Alec Radford, Jeffrey Wu, Rewon Child, David Luan, Dario Amodei, Ilya
  Sutskever, et~al. 2019.
\newblock Language models are unsupervised multitask learners.
\newblock \emph{OpenAI blog}, 1(8):9.

\bibitem[{Raffel et~al.(2020)Raffel, Shazeer, Roberts, Lee, Narang, Matena,
  Zhou, Li, Liu et~al.}]{raffel2020exploring}
Colin Raffel, Noam Shazeer, Adam Roberts, Katherine Lee, Sharan Narang, Michael
  Matena, Yanqi Zhou, Wei Li, Peter~J Liu, et~al. 2020.
\newblock Exploring the limits of transfer learning with a unified text-to-text
  transformer.
\newblock \emph{J. Mach. Learn. Res.}, 21(140):1--67.

\bibitem[{Rebuffi et~al.(2017)Rebuffi, Bilen, and Vedaldi}]{Rebuffi:17}
Sylvestre{-}Alvise Rebuffi, Hakan Bilen, and Andrea Vedaldi. 2017.
\newblock \href {http://arxiv.org/abs/1705.08045} {Learning multiple visual
  domains with residual adapters}.
\newblock \emph{CoRR}, abs/1705.08045.

\bibitem[{Schick and Sch{\"u}tze(2020)}]{schick2020exploiting}
Timo Schick and Hinrich Sch{\"u}tze. 2020.
\newblock Exploiting cloze questions for few shot text classification and
  natural language inference.
\newblock \emph{arXiv preprint arXiv:2001.07676}.

\bibitem[{Shin et~al.(2020)Shin, Razeghi, Logan~IV, Wallace, and
  Singh}]{shin2020autoprompt}
Taylor Shin, Yasaman Razeghi, Robert~L Logan~IV, Eric Wallace, and Sameer
  Singh. 2020.
\newblock Autoprompt: Eliciting knowledge from language models with
  automatically generated prompts.
\newblock \emph{arXiv preprint arXiv:2010.15980}.

\bibitem[{Su et~al.(2021)Su, Shu, Mansimov, Gupta, Cai, Lai, and
  Zhang}]{su2021multi}
Yixuan Su, Lei Shu, Elman Mansimov, Arshit Gupta, Deng Cai, Yi-An Lai, and
  Yi~Zhang. 2021.
\newblock Multi-task pre-training for plug-and-play task-oriented dialogue
  system.
\newblock \emph{arXiv preprint arXiv:2109.14739}.

\bibitem[{Sutskever et~al.(2014)Sutskever, Vinyals, and Le}]{sutskever-s2s}
Ilya Sutskever, Oriol Vinyals, and Quoc~V. Le. 2014.
\newblock Sequence to sequence learning with neural networks.
\newblock In \emph{Proceedings of the 27th International Conference on Neural
  Information Processing Systems - Volume 2}, NIPS'14, page 3104–3112,
  Cambridge, MA, USA. MIT Press.

\bibitem[{Wen et~al.(2015)Wen, Gasic, Mrksic, Su, Vandyke, and
  Young}]{wen2015semantically}
Tsung-Hsien Wen, Milica Gasic, Nikola Mrksic, Pei-Hao Su, David Vandyke, and
  Steve Young. 2015.
\newblock Semantically conditioned lstm-based natural language generation for
  spoken dialogue systems.
\newblock \emph{arXiv preprint arXiv:1508.01745}.

\bibitem[{Yang et~al.(2021)Yang, Li, and Quan}]{yang2021ubar}
Yunyi Yang, Yunhao Li, and Xiaojun Quan. 2021.
\newblock Ubar: Towards fully end-to-end task-oriented dialog system with
  gpt-2.
\newblock In \emph{Proceedings of the AAAI Conference on Artificial
  Intelligence}, volume~35, pages 14230--14238.

\bibitem[{Zang et~al.(2020)Zang, Rastogi, Sunkara, Gupta, Zhang, and
  Chen}]{zang2020multiwoz}
Xiaoxue Zang, Abhinav Rastogi, Srinivas Sunkara, Raghav Gupta, Jianguo Zhang,
  and Jindong Chen. 2020.
\newblock Multiwoz 2.2: A dialogue dataset with additional annotation
  corrections and state tracking baselines.
\newblock In \emph{Proceedings of the 2nd Workshop on Natural Language
  Processing for Conversational AI, ACL 2020}, pages 109--117.

\bibitem[{Zhu et~al.(2022)Zhu, Li, Mi, Zhu, and Huang}]{zhu2022continual}
Qi~Zhu, Bing Li, Fei Mi, Xiaoyan Zhu, and Minlie Huang. 2022.
\newblock Continual prompt tuning for dialog state tracking.
\newblock \emph{arXiv preprint arXiv:2203.06654}.

\end{thebibliography}
\bibliographystyle{acl_natbib}

\appendix
\section{Human Evaluation Task}
\label{appendix:ex}

We explored contextual dynamic prompting strategies for the response generation task using the MultiWOZ 2.2 \cite{budzianowski2018large, zang2020multiwoz} dataset and noticed that the combined score that we obtained was significantly better than the baseline prefix-tuning method of response generation. To understand if the agents which incorporated contextual dynamic prompts did indeed provide a better conversational experience, we designed a small human evaluation task to test our hypothesis.\\

We picked a random subset of 10\% of the conversations from the original MultiWOZ test data to perform this analysis. Once we obtained this random set, we ran our four model variants as described in Section \ref{Conxtual dynamic prompting} on the conversations to obtain system responses for each of them. We then presented the different agents' responses to the annotator as shown in Table \ref{tab:annot_table} below. In order to avoid potential biases, we shuffled the order of the agents between our annotators i.e., Agent 1 for annotator \textit{a} would not be Agent 1 for annotator \textit{b}. We kept track of which agents corresponded to which of our four methods prior to distribution of data amongst the annotators.\\

The annotators were given instructions to read every turn of conversation and provide a number between 1 and 4 for the agent which they thought performed the best for that turn. If the annotators found that there was a tie, they could pick more than one agent as [agent\_a, agent\_b]. In addition to this instruction, annotators were asked to read the entire conversation and pick the agent which performed the best - once again with an option to pick multiple. Table \ref{tab:annot_style} below shows an example annotation style for a single conversation spanning 6 turns. There is an annotation at every turn and a single annotation at the end of the conversation.\\

We tallied results and re-mapped all agents back to their methods and found that agents 3 and 4 as mentioned in Section \ref{section:human_eval} were preferred at the conversation level in a total of 53 of the 100 conversations while agents 1 and 2 were only preferred 10 conversations in the entire set of 100. 

\begin{table*}[t]
  \centering
  \begin{tabular}{|c|c|c|c|c|c|}
    \hline
    Turn num & User turn & Agent 1 response & Agent 2 response & Agent 3 response & Agent 4 response \\
    \hline
    1 & & & & & \\
    2 & & & & & \\
    3& & & & & \\
    4& & & & & \\
    5& & & & & \\
    6& & & & & \\
    7& & & & & \\
    \hline
  \end{tabular}
  \caption{The format which is presented to annotators while performing turn-level and conversation-level annotation. The agents are shuffled between the annotators to avoid biasing them.}
  \label{tab:annot_table}
\end{table*}

\begin{table*}[t]
  \centering
  \begin{tabular}{|c|c|c|}
    \hline
    Turn num & Turn level & Conversation level \\
    \hline
    1& 2 & \\
    2& [3,4] & \\
    3& 2& \\
    4& 3 & \\
    5& 4 & \\
    6& [3,4]& 3\\
    \hline
  \end{tabular}
  \caption{We asked annotators to provide two levels of annotation for each conversation - turn level and conversation level. Annotators were allowed to pick more than agent in case of a tie.}
  \label{tab:annot_style}
\end{table*}

\section{Examples}
\label{appendix:eval}

In the table below, we provide another examples from the human annotation task where annotators picked agent 4 as the best performing agent for the conversation (we performed the re-mapping to fit the agent numbers in Section \ref{section:human_eval}). Table \ref{tab:example_1} shown in the main paper indicates a scenario where Agent 4 was rightly able to ask the customer about their area of preference for dining, and also considers a booking time change from the user. Table \ref{tab:example_2} shows an example where agent 4 is able to take into consideration all the different constraints mentioned by the user in terms of slots in the response that it generates. Table \ref{tab:example_3} shows an example where the agent 4 is able to correctly respond to the user request for number of people to reserve the table for - which none of the other agents captured. All these examples showed that agent 4, which uses contextual dynamic prompts with dialog states was indeed able to provide smoother responses to the customer input compared to the other agents.

\begin{table*}[t]
    \small
  \centering
  \begin{tabular}{|c|p{25mm}|p{25mm}|p{25mm}|p{25mm}|p{25mm}|}
    \hline
    Turn num & User turn & Agent 1 response & Agent 2 response & Agent 3 response & Agent 4 response \\
    \hline
    1 & i am looking for a train that departs from peterborough and goes to cambridge.  & there are [choice] trains leaving from [departure] to [destination]. what day would you like to travel? & what day would you like to travel? & there are [choice] trains leaving [departure] from [departure] to [destination]. what day would you like to travel? & i have [choice] trains departing from [departure] to [destination]. what day would you like to travel?\\
    \hline
    2 & we will be leaving on sunday afternoon.  & [trainid] leaves [departure] at [leaveat] and arrives in [destination] at [arriveby]. would you like me to book that for you? & i have [choice] trains available. what time would you like to leave?& [trainid] leaves at [leaveat] and arrives in [destination] at [arriveby]. would you like me to book it for you?& [trainid] leaves [departure] at [leaveat] and arrives at [destination] at [arriveby]. would that work for you?\\
    \hline
    3& yes, it should leave after 20:15 and leave on monday, not sunday. please give me a price, travel time, and arrival time for any train.  & [trainid] leaves at [leaveat] and arrives in [destination] at [arriveby]. the price is [price]. & [trainid] leaves at [leaveat] and arrives at [arriveby]. would you like me to book that for you? & [trainid] leaves at [leaveat] and arrives at [arriveby]. the price is [price]. would you like me to book it for you? & [trainid] leaves [departure] at [leaveat] and arrives in [destination] at [arriveby]. the cost is [price] and the travel time is [duration]. \\
    \hline
    4& yes please. what is the travel time? i'll also need a hotel in the expensive price range. can you help me with that as well?  & the travel time is [duration]. is there anything else i can help you with? & the travel time is [duration]. & the travel time is [duration]. is there anything else i can help you with? & the travel time is [duration]. is there anything else i can help you with?\\
    \hline
    5& yes, i'd like a 2 star rating and i'll need it to be in the east side of town.  & i have [choice] [type] in the [area] with [stars] stars. do you have a preference? & there are [choice] [type] in the [area] that are [stars] stars. do you have a preference? & there are [choice] [type] in the [area] that are [pricerange] and [pricerange]. do you have a preference?& [name] is a [stars] star hotel in the [area]. would you like me to book it for you? \\
    \hline
    6& yes, i would like to book it for 3 people starting that same day for 4 nights.  & booking was successful. reference number is : [ref]. is there anything else i can help you with? & booking was successful. reference number is : [ref]. & booking was successful, the total fee is [price] payable at the station. reference number is : [ref].& booking was successful. the total fee is [price] payable at the station. reference number is : [ref]. \\
    \hline
    7& that is all thank you goodbye.  & you're welcome. have a great day! & you're welcome. have a great day. & you're welcome. have a great day.& you're welcome. have a great day. \\
    \hline
  \end{tabular}
  \caption{Another example where Agent 4 was picked as the best performing agent in a conversation.}
  \label{tab:example_2}
\end{table*}

\section{Compute Infrastructure and Experimental settings used}
All our experiments were carried out on public AWS p3 GPU instances which are backed by Tesla V100 GPU cores. All four variants of our models required 18-24 hours to reach convergence on these machines. Learning rates were varied based on performance of models on the validation split of our dataset.

\end{document}